\documentclass[sigconf]{acmart}
\AtBeginDocument{%
  \providecommand\BibTeX{{%
    \normalfont B\kern-0.5em{\scshape i\kern-0.25em b}\kern-0.8em\TeX}}}

\setcopyright{rightsretained}
\copyrightyear{2023}
\acmYear{2023}
\acmDOI{}

\acmConference[arXiv '23]{arXiv Preprint}{2023}{arXiv}
\acmBooktitle{arXiv '23: arXiv Preprint,
2023, Online} 
\acmPrice{}
\acmISBN{}

\usepackage{pdfpages}
\usepackage{bm}
\usepackage{float}
\usepackage{subcaption}
\usepackage{rotating}
\usepackage{appendix}
\usepackage{threeparttable}
\usepackage{booktabs,threeparttable}
\usepackage{multirow}
\usepackage{mathtools}
\usepackage{bbm}
\usepackage{pdflscape}
\usepackage{enumitem}

\def\kdd{0}

\begin{document}

\title{Inclusive FinTech Lending via Contrastive Learning and Domain Adaptation}


\author{Xiyang Hu}
\affiliation{%
  \institution{Carnegie Mellon University}
  \city{Pittsburgh}
  \state{PA}
  \country{USA}
  \postcode{15213}}
\email{xiyanghu@cmu.edu}

\author{Yan Huang}
\affiliation{%
  \institution{Carnegie Mellon University}
  \city{Pittsburgh}
  \state{PA}
  \country{USA}
  \postcode{15213}}
\email{yanhuang@cmu.edu}

\author{Beibei Li}
\affiliation{%
  \institution{Carnegie Mellon University}
  \city{Pittsburgh}
  \state{PA}
  \country{USA}
  \postcode{15213}}
\email{beibeili@andrew.cmu.edu}

\author{Tian Lu}
\affiliation{%
  \institution{Arizona State University}
  \city{Tempe}
  \state{AZ}
  \country{USA}
  \postcode{85287}}
\email{lutian@asu.edu}

\renewcommand{\shortauthors}{Hu, et al.}

\begin{abstract}
  FinTech lending (e.g., micro-lending) has played a significant role in facilitating financial inclusion. 
  However, there are concerns about the potentially biased algorithmic decision-making during loan screening. Machine learning algorithms used to evaluate credit quality can be influenced by representation bias in the training data, as we only have access to the default outcome labels of approved loan applications, for which the borrowers' socioeconomic characteristics are better than those of rejected ones. In this case, the model trained on the labeled data performs well on the historically approved population, but does not generalize well to borrowers of low socioeconomic background. 
  In this paper, we investigate the problem of representation bias in loan screening for a real-world FinTech lending platform. 
  We propose a new Transformer-based sequential loan screening model with self-supervised contrastive learning and domain adaptation to tackle this challenging issue. 
  We use contrastive learning to train our feature extractor on unapproved (unlabeled) loan applications and use domain adaptation to generalize the performance of our label predictor. We demonstrate the effectiveness of our model through extensive experimentation in the real-world micro-lending setting. Our results show that our model significantly promotes the inclusiveness of funding decisions, while also improving loan screening accuracy and profit by 7.10\% and 8.95\% respectively. We also show that incorporating the test data into contrastive learning and domain adaptation and labeling a small ratio of test data can further boost model performance.
\end{abstract}

\ccsdesc[500]{Computing methodologies~Neural networks}
\ccsdesc[500]{Information systems~Data mining}
\ccsdesc[500]{Applied computing~Electronic commerce}
\ccsdesc[500]{Applied computing~Enterprise computing}

\keywords{Financial Inclusion, Representation Bias, Contrastive Learning, Domain Adaptation, Fairness, FinTech, Credit Risk Prediction}



\maketitle

\section{Introduction}
\if\kdd
Financial inclusion has been the essential focus of worldwide policymakers because of its substantial impacts on economic growth and development. It refers to the accessibility of useful and affordable financial products and services for individuals and businesses \citep{wordbank2021fin}. This includes things like bank accounts, credit, insurance, and payment systems. It contributes to poverty elimination, inequality reduction, job growth, and financial stability \citep{demirgucc2017financial, morgan2018financial, imf2020the, un2021fin}. 
Financial institutions have incentives to advance financial inclusion not only because they want to make a positive societal impact, but also because providing financial products and services to the underserved market can help them make additional profits \citep{davis2021driving}.
Financial inclusion issues are concerning in both developing and developed countries \citep{beck2008banking, fungavcova2015understanding, bogan2022intersectionality}. A recent national survey in the United States indicates that 4.5 percent of U.S. households were “unbanked”, and 14.1 percent of U.S. households were “underbanked” \citep{dfic2021survey}.

Financial exclusion refers to the inability or difficulty of individuals or communities to access financial services such as banking, credit, and insurance. This can have significant negative impacts on individuals and communities, as it can limit their ability to save and invest, access credit, and protect themselves against financial risks. Financial exclusion can be caused by a variety of factors, including lack of education and financial literacy, discrimination, and high transaction costs. It is often disproportionately experienced by marginalized and vulnerable groups, such as low-income communities and individuals, racial and ethnic minorities, and rural populations. The representation bias in the dataset discussed earlier can also contribute to financial exclusion, as it may exclude certain groups from being able to access financial services. It is therefore important to address representation bias in order to promote financial inclusion and ensure that all individuals and communities have equal access to financial services.
\fi

FinTech innovations over the years have been a driving force in facilitating financial inclusion. It has changed the way financial institutions create and deliver products and services, and the way how they offer customers democratized access to financial services \citep{philippon2019FinTech}. 
\if\kdd
This has been particularly beneficial for underserved communities, such as those living in rural areas or those with low incomes, who may have previously had difficulty accessing traditional financial services.
According to a report by McKinsey \& Company, FinTech could provide financial service access to 1.6 billion unbanked people, create 2.1 trillion new loans to individuals and small businesses, and increase the GDPs of all emerging economies by 6\% by 2025 --- a total of \$3.7 trillion \citep{mckinsey2016digital}. Examples of transformative FinTech innovations to date include mobile payment systems, new digital advisory and trading systems, crowdfunding platforms, online lending, machine learning and artificial intelligence, etc.
\fi
For example, micro-lending creates a more inclusive financial system by lowering processing time and operational costs, improving the user experience, and more importantly, hoping to grant loans to borrowers who are not able to receive credit from traditional lenders \citep{berg2022FinTech, cornelli2022impact}. It is inevitably challenging to assess the credit quality of a wider pool of candidates because those traditionally under-served candidates usually do not have sufficient credit history and present distinct characteristics or behavioral patterns from those "regular" candidates who have sufficient credit records \citep{lu2023profit}. Many micro-lending companies, therefore, turn to machine learning techniques to improve the effectiveness and efficiency of borrower screening. 
However, these credit assessing algorithms are often found to favor borrowers of specific gender, race, occupation, income, etc.
\citep{fu2021crowds, bartlett2022consumer, so2022beyond} and thus, preventing people with disadvantaged socioeconomic backgrounds from being served.

One important source of this problem is the \textit{representation bias} --- some parts of the population are underrepresented by the training samples, and thus the trained model fails to generalize well to this subset of the population \citep{suresh2021framework}. The problem of unbalanced representation can be exacerbated by the \textit{selective labels} problem. In the micro-lending setting, we only have the default outcome labels for approved loan applications but not for those that were denied, and traditional practice only uses these samples to train the machine learning model. In addition, usually, the size of these approved loan applications is much smaller than that of rejected ones.
These, together, make the training dataset skewed towards the historically approved borrowers, who overall have more favorable socioeconomic characteristics. 
This historically approved subset for training cannot reflect the distribution of the whole applicant population, and therefore renders the algorithms favor applicants who have similar characteristics to those traditionally thought "good borrowers" \citep{cowgill2019economics}, impeding those with less favorable socioeconomic backgrounds from accessing credits.

Recent advances in \textit{self-supervised} machine learning, which does not require any labels for model training, provide new possibilities to deal with the representation bias. 
Self-supervised learning 
is a subset of unsupervised learning where some kind of supervisory signal is generated automatically from the unlabeled dataset. Therefore, in the micro-lending setting, we can leverage the unlabeled/unapproved loan applications to train credit-scoring machine learning models to improve financial inclusion. 
Specifically, in this work, we use self-supervised contrastive learning to train our feature extractor on unapproved loan application samples. The intuition of the contrastive learning loss function is to minimize the distance between positive samples while maximizing the distance between negative samples \citep{chen2020simple}. We generate positive pair examples through data augmentation. Given a positive pair, we use all other augmented examples within the same batch as negative examples. 

In addition to using self-supervised contrastive learning to train our feature extractor on the unlabeled/unapproved loan application records, we also incorporate domain adaptation techniques to solve the distribution shift problem between labeled loans and unlabeled loans, and between training samples and test samples. This is because the contrastive learning loss only optimizes the feature extractor to learn effective representation from unlabeled samples, but the label predictor is still only trained on labeled samples. Incorporating domain adaptation can generalize the performance of the label predictor, which is trained to fit the distribution of the labeled training loans, to the unlabeled loans and the test samples. 

We train and test our model in a real-world loan screening setting of a leading micro-loan platform. Note that our test dataset was collected from an experiment conducted on the platform, during which period all loan applications were approved \textit{without} any screening. This unique dataset contains the ground truth labels of the \textit{entire} borrower pool. We evaluate our model's decision outcome on this test dataset, which does not have any representation bias and therefore guarantees our evaluations reflect the model's real-world performance.
Our analysis indicates that our model significantly improves loan prediction performance and expands loan access to broader low-socioeconomic borrowers. We summarize our contributions as follows:
\begin{itemize}[noitemsep,topsep=2pt,parsep=2pt,partopsep=0pt, leftmargin=0.14in]
    
    \item We consider the selective label problem in micro-lending, where we only have labels for historically approved loan applications. These historically approved borrowers overall have more favorable socioeconomic characteristics. This makes traditional supervised loan screening algorithms trained on the labeled data tend to underserve people from disadvantaged backgrounds.
    \item We propose a sequential loan screening model, which incorporates contrastive learning and domain adaptation to effectively learn from unlabeled samples. Extensive experiments demonstrate that our model outperforms baseline models and improves financial inclusion. We also conduct in-depth explorations on further boosting model performance by introducing test samples into self-supervised training, and by labeling a small ratio of test samples.
\end{itemize}

\section{Research Context and Dataset}

Our research context is the loan screening task in FinTech lending. Our model training and evaluation are based on a real-world setting of a micro-lending platform in Asia. The platform offers microloans averaging around \$450 USD using its own funds (it does not involve external lenders). 
\if\kdd
These loan applicants typically borrow money on this platform to meet temporary financial needs, such as supplementary working capital for small businesses, education expenses, medical bills, or irregular shopping needs. 
\fi
To apply, applicants must provide their personal information and demographics, including their name, age, gender, education, housing, income level, etc. They are also required to list 3 to 4 contact people (who must be family or close friends). The loan term typically ranges from 3 to 8 months and the platform charges an annual interest rate of about 18\%. 

During the period covered by our training dataset, the platform manually evaluated applicants' creditworthiness using its human employees (evaluators), rather than using machine learning algorithms. The evaluators were trained regularly to maintain consistency in their evaluation criteria, which were based on their collective daily work experience. The platform did not train its evaluators on issues of fairness or inclusion in credit risk evaluation. The loan application process involves randomly assigning loan applications to an evaluator, who then evaluates the provided information and decides whether to approve or reject the loan application. The loan approval decision is based on whether the loan is expected to go into default or not. The loan is repaid in monthly installments starting one month after the loan is issued. 
According to the platform's rules, default occurs when a loan is unpaid for 90 days or more after the due date. The default probabilities for new applicants are based on the personal information provided, while repeat applicants who have previously received loans from the platform are also evaluated based on their repayment performance on previous loans. Specifically, the repayment performance is captured by three metrics: (1) the number of overdue days, (2) the proportion of installments for which the borrower showed a positive attitude towards repayment, and (3) the proportion of installments that were repaid with financial assistance from family or friends. These three performance signals are collected from records of interactions between the platform and borrowers during the repayment process. They reflect different aspects of borrowers' creditworthiness and reliability.

\subsection{Dataset}

The training dataset consists of longitudinal loan records spanning 33 months. During this period, there were 311,200 loan applications, of which 135,938 were approved (approval rate 43.68\%) and 175,262 were rejected. The sample includes 139,455 applicants, with an average of 2.23 loan applications per applicant. 
38.37\% of applicants (53,503) have applied more than once (referred to as "repeat applicants"), while the rest have only applied once ("single-time applicants"). Repeat applicants had an average of 4.21 loan applications and an approval rate of 49.45\% for their second and subsequent applications, while the approval rate for the first application of all applicants was 36.58\%. 

The dataset includes demographic and socioeconomic information of each applicant, such as education level, monthly income, disposable personal income per capita, and house ownership. It also contains loan information such as loan amount and term, as well as repayment information for approved loans, including the loan defaulting or not, the profit or the loss of the loan, and three repayment performance signals for the loan. 

\begin{figure}[t]
\begin{center}
\includegraphics[width=0.47\textwidth]{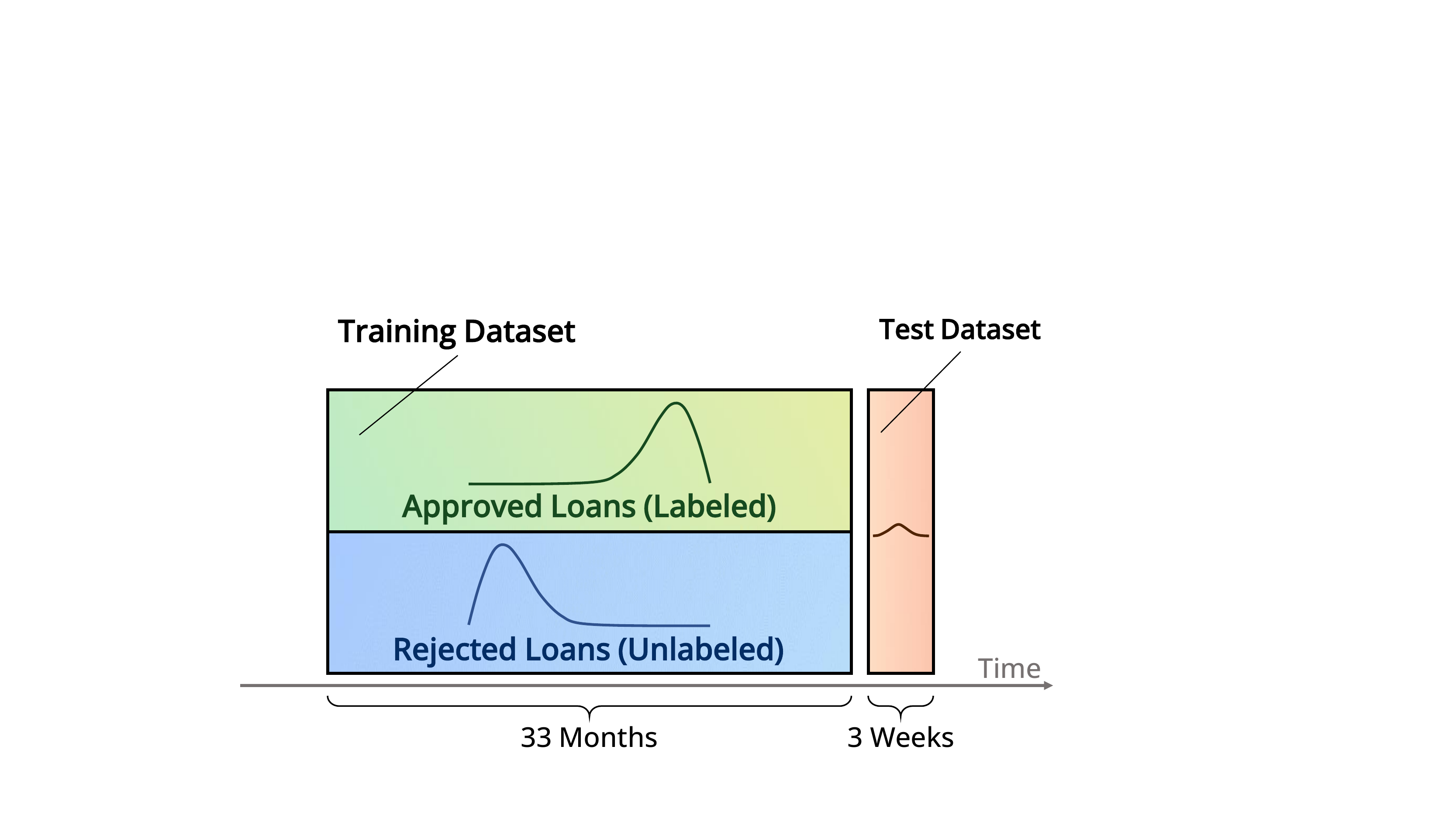}
\vskip -2mm
\caption{Dataset. Traditional supervised loan screening algorithms, e.g. XGBoost, only use historically approved loans as the training dataset. We incorporate the samples of rejected loans into training through self-supervised contrastive learning and domain adaptation. For our test dataset, we got their labels by approving all these loan applications.} 
\label{fig:data}
\end{center}
\end{figure}

In order to evaluate our model's performance, we have to obtain the true labels of all evaluation samples. We conducted a three-week experiment with the FinTech lending platform, for which all loan applications were approved without any screening. This experiment allowed us to observe the loan repayment behaviors of all applicants on the platform without any selection bias for a period of time. All loan records within this three-week experimental period are used as the evaluation dataset, which includes a total of 5,999 loans. This experiment made it possible for us to quantify our model's prediction performance in the real world, which has been difficult for previous studies to do due to the lack of true labels for loans that were not approved in reality. Figure~\ref{fig:data} illustrates the datasets involved in this paper.

\begin{table}[t]
\centering
\footnotesize
\begin{tabular}{llrr}
\toprule
   &     \textbf{Feature}    & \multicolumn{1}{l}{\textbf{Approved loan}} & \multicolumn{1}{l}{\textbf{Rejected loan}} \\ \midrule
\multirow{4}{*}{\begin{tabular}[c]{@{}l@{}}The means of \\ the feature of \\ unique \\ borrowers\end{tabular}} & Living-city DPI      & 43760.42   & 36790.70     \\
& Monthly income level & 4.03   & 2.90  \\
  & Education level   & 2.53    & 2.10   \\& Homeownership       & 0.22   & 0.13 \\ \bottomrule
\end{tabular}
\caption{Comparison of the characteristics of the approved/labeled loans and rejected/unlabeled loans}
\label{tab:comp-loan-appr}
\vskip -6mm
\end{table}

Table~\ref{tab:comp-loan-appr} shows the comparison of the characteristics between the borrowers whose loan applications got approved and those whose got rejected.
It is clear that compared with the loan-rejected borrowers, the borrowers whose loan applications were approved in the dataset live in a better-developed city, have higher income, possess a higher degree, and own better housing. This indicates if we use data on just the approved loans to train an algorithm, there will be a representation bias in the labeled training dataset. The labeled training data does not accurately represent the entire population, especially the low socioeconomic people. Training on historically approved loans would ignore the rejected ones who are actually able to repay the loan but from lower socioeconomic backgrounds. This bias can have significant implications for any downstream machine learning model training. In order to obtain a model that is financially inclusive, it is important to carefully consider and account for the representation bias in the dataset.

\section{Model}

Given a borrower $i$, we have her demographics $\mathbf{D}_i$ and her loan application records $\mathbf{A}_i$. $\mathbf{A}_i$ is a sequence of loan applications $\mathbf{A}_i=[\mathbf{a}_{i1} ~\cdots~ \mathbf{a}_{it} ~\cdots~  \mathbf{a}_{iT_i} ]^\mathsf{T}  \in \mathbb{R}^{T_i \times 3}$. Each loan application record $\mathbf{a}_{it}$ is a vector containing three scalars: (1) loan amount, (2) loan interest rate, and (3) loan term. For each loan application $\mathbf{a}_{it}$, if it was approved, we have the records of the borrower's repayment behavior on this loan, and we also observe the label on this loan's default outcome. Similarly, we use a sequence of the same length with $\mathbf{A}_i$ to denote the repayment behavior sequence of user $i$ as $\mathbf{R}_i=[\mathbf{r}_{i1} ~\cdots~ \mathbf{r}_{it} ~\cdots~  \mathbf{r}_{i|\mathbf{R}_i|}]^\mathsf{T}  \in \mathbb{R}^{T_i \times 3}$, where $\mathbf{r}_{it}$ is the repayment behavior of loan $\mathbf{a}_{i,t-1}$. Each $\mathbf{r}_{it}$ is a vector of three scalars: (1) the number of overdue days, (2) the proportion of installments for which the borrower showed a positive attitude towards repayment, and (3) the proportion of installments that were repaid with financial assistance from family or friends. These repayment performance signals are extracted from records of interactions between the platform and borrowers during the repayment process, and provide insight into the borrowers' creditworthiness. For unapproved loans, all the three repayment behavior values are filled with 0.
We shift the repayment behavior sequence $\mathbf{R}_i$ by one time unit for computational convenience --- when we do the loan screening for $\mathbf{a}_{it}$, we are using the loan repayment behavior of previous loans $\{\mathbf{a}_{i1}, \cdots,  \mathbf{a}_{i,t-1}\}$.

If a loan application $\mathbf{a}_{it}$ was approved, we observe its label $Y_{it}$. $Y_{it}=1$ indicates loan $\mathbf{a}_{it}$ was approved and non-default, $Y_{it}=0$ indicates it was approved but default, and we use $Y_{it}=-1$ to indicate this loan application was not approved and its label is unobserved. For convenience, we use $S_{it} \coloneqq \mathbbm{1}_{[ Y_{i,t-1}\neq-1 ]}$ to indicate whether we have records for the loan repayment behavior of $\mathbf{a}_{i,t-1}$. That is, if loan $\mathbf{a}_{i,t-1}$ was approved, then we observe its repayment behavior. 

\begin{figure*}[t]
\begin{center}
\includegraphics[width=\textwidth]{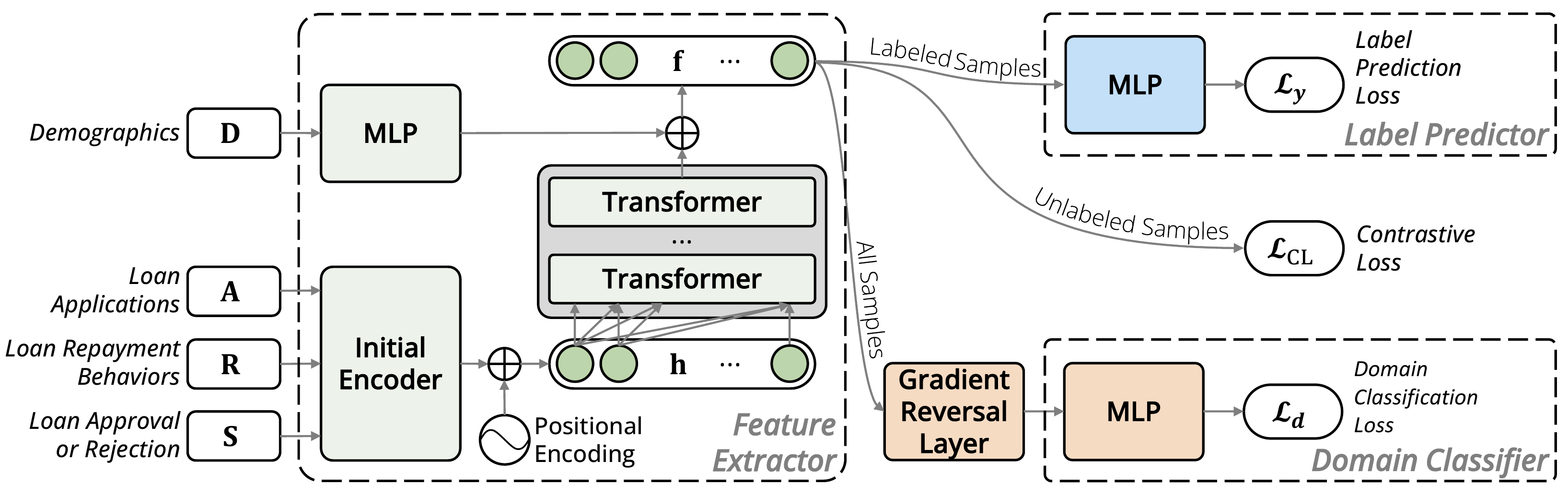}
\vskip -2mm
\caption{Model Architecture} 
\label{fig:model}
\end{center}
\end{figure*}

Figure~\ref{fig:model} shows our model framework. We first use an initial encoder to map the loan records into the initial embedding space. Then we add a positional encoding to inject information about the relative or absolute position of each loan application in the sequence of this applicant. After this, we use a transformer-based sequence encoder to encode the loan sequence. In addition to the sequential loan records, we also have access to the applicants' demographics. We use a simple multilayer perceptron (MLP) to map the demographic information into the same space. The final feature vector/embedding $\mathbf{f}$ is a fusion of the encoded demographic information and the encoded loan sequence information, which is achieved through element-wise addition. The feature vectors are then fed into three modules: 
\begin{itemize}[noitemsep,topsep=2pt,parsep=2pt,partopsep=2pt, leftmargin=0.14in]
    \item The label predictor module calculates the label prediction cross-entropy loss $\mathcal{L}_y$ and outputs the predicted labels. It only trains on labeled (approved) loans. Minimizing the label prediction loss $\mathcal{L}_y$ optimizes the feature extractor and the label predictor to be discriminative on whether a loan would default or not.
    \item The contrastive learning module calculates the contrastive loss $\mathcal{L}_\text{CL}$ on unlabeled (unapproved) loans. Minimizing the contrastive loss $\mathcal{L}_\text{CL}$ helps the feature extractor to learn feature vectors with meaningful semantics in a self-supervised way.
    \item The domain classifier module predicts a loan is from the labeled domain or the unlabeled domain. It calculates the domain classification cross-entropy loss $\mathcal{L}_d$ on all samples, i.e. both labeled loans and unlabeled loans. It is used together with a gradient reversal layer. Minimizing the contrastive loss $\mathcal{L}_d$ pushes the feature extractor to learn loan representations so that the distribution of labeled and unlabeled loan representations are more similar to each other. This helps the label predictor to generalize its classification ability to unlabeled loans.
\end{itemize}

The training of the whole model is based on a weighted sum of the three losses:
\begin{equation}
    \mathcal{L} = w_y\mathcal{L}_y + w_\text{CL}\mathcal{L}_\text{CL} + w_d\mathcal{L}_d
\end{equation}
where $w_y$, $w_\text{CL}$ and $w_d$ are their corresponding weights. In our training, we normalize $w_y$ to be 1. We set $w_\text{CL}$ to be 0.1, and $w_d$ to be $0.1 \cdot (\frac{2}{1+\exp(-\gamma \cdot p)}-1)$, where $\gamma$ is a hyperparameter set to be 0.001 and $p$ is the training step --- we gradually increase $w_d$ from 0 to 0.1 to avoid noisy signal at the early stages of the training the domain classifier.

\subsection{Initial Encoder}

We first concatenate the loan application features and the loan repayment behavior features, and denote it as $\mathbf{C}_i \coloneqq [\mathbf{A}_i~\mathbf{R}_i]  \in \mathbb{R}^{T_i \times 6}$. Then we do a linear transformation on $\mathbf{C}_i$ to map it into a space of dimension size 64. 
\begin{equation}
\small
\mathbf{h} = (1-\mathbf{S}_i) \odot (\mathbf{C}_i \mathbf{W}_0 + \mathbf{b}_0) + \mathbf{S}_i \odot (\mathbf{C}_i \mathbf{W}_1 + \mathbf{b}_1) + \text{positional encoding}
\end{equation}
where $\mathbf{W}_0, \mathbf{W}_1 \in \mathbb{R}^{6 \times 64}$ and $\mathbf{b}_0, \mathbf{b}_1 \in \mathbb{R}^{1 \times 64}$ are the parameters of the linear layers, and $\odot$ is the element-wise multiplication operation. The intuition is that we have two different linear transformation heads according to whether we observe the repayment behavior of previous loans: $(\mathbf{W}_0, \mathbf{b}_0)$ is for the loan applications whose $S_{it}$ is 0, and $(\mathbf{W}_1, \mathbf{b}_1)$ is for the loan applications whose $S_{it}$ is 1. We also add a positional encoding to incorporate the position information of each element in the sequence. The positional encoding is realized through \texttt{torch.nn.Embedding}, which is basically a look-up table whose parameters get updated during training.



\subsection{Transformer}

Transformer is a type of neural network architecture for sequence modeling that was introduced by \cite{vaswani2017attention}. It has been widely used in natural language processing tasks such as language modeling, machine translation, and text summarization because of its ability to handle long-range dependencies in sequences effectively. The key innovation of the Transformer architecture is the use of attention mechanisms, which allow the model to focus on different parts of the input sequence at different times. This makes it well-suited for tasks that require processing sequences with variable lengths and for handling long-range dependencies, as the model can attend to relevant parts of the input sequence at different steps in the computation. 
\if\kdd
Transformer models have achieved state-of-the-art results on a wide range of NLP tasks, and have become one of the most widely used architectures in the field. They have also been applied to other domains such as computer vision, audio processing, recommendation, etc.
\fi

An attention mechanism uses a query and a set of key-value pairs to generate an output vector, where the output is a combination of the values, with the weight of each value determined by a function of the query and its corresponding key. All of the query, keys, values, and output are vectors.

We use a transformer architecture to encode the loan application sequences into feature embeddings. The input to the transformer encoder is the initial embedding $\mathbf{h}$ as we noted above. We first use three projection matrices  $\mathbf{W}_Q$, $\mathbf{W}_K$, $\mathbf{W}_V \in \mathbb{R}^{64 \times 64}$ to map $\mathbf{h}$ into three matrix query $\mathbf{Q}$, key $\mathbf{K}$, and value $\mathbf{V}$: 
\begin{equation}
    \mathbf{Q} = \mathbf{h} \mathbf{W}_Q,~\mathbf{K} = \mathbf{h} \mathbf{W}_K,~\mathbf{V} = \mathbf{h} \mathbf{W}_V
\end{equation}

Then we use the attention mechanism to do the computation in the following way:
\begin{equation}
    \texttt{Attention}(\mathbf{Q}, \mathbf{K}, \mathbf{V}) = \texttt{softmax}(\frac{\mathbf{Q}\mathbf{K}^\mathsf{T}}{\sqrt{d_k}})\mathbf{V}
\end{equation}
where $d_k$ is the dimension $\mathbf{K}$. The purpose of the scaling factor $\frac{\mathbf{Q}\mathbf{K}^\mathsf{T}}{\sqrt{d_k}}$ is to rescale the variance of $\frac{\mathbf{Q}\mathbf{K}^\mathsf{T}}{\sqrt{d_k}}$ to be one. Therefore, the attention function outputs the weighted sum of the values $V$, where the weights are $\texttt{softmax}(\frac{\mathbf{Q}\mathbf{K}^\mathsf{T}}{\sqrt{d_k}})$.

\if\kdd
\fi

The output of the attention layer is fed into an \texttt{Add \& Norm} module, a \texttt{Feed Forward} module, and an \texttt{Add \& Norm} module again. The \texttt{Add \& Norm} module conducts a pointwise addition of the attention's input and output, i.e. $\mathbf{h}$ and $\texttt{Attention}(\mathbf{Q}, \mathbf{K}, \mathbf{V})$, and a layer normalization is done afterwards. The \texttt{Feed Forward} module is a two-layer perceptron with a non-linear activation function in-between the two linear layers, and a layer normalization and a dropout operation afterwards. Dropout \citep{srivastava2014dropout} is a regularization technique for neural networks that helps to prevent overfitting. It works by randomly setting a fraction of the dropout input units to zero during the training process. 
\if\kdd
This has the effect of reducing the complexity of the model, as the network is forced to rely on the remaining units to make predictions.
\fi
In our model, we stack two transformers together to encode the loan sequence into feature vectors $\mathbf{f}_A$.

\subsection{Multimodal Fusion}

\if\kdd
Multimodal fusion refers to the process of combining information from multiple modalities, or sources of data. For example, multimodal fusion might be used to combine information from tabular data, time series sequences, text, images, and audio to create a more complete understanding of a situation or to make a decision. Late fusion is a kind of multimodal fusion method that combines the results of multiple models or systems after they have already been independently processed. 
\fi

Note that the transformer module only takes a loan application sequence as the input, while the applicant's demographic information remains unused. We adopt the late fusion method of multimodal machine learning \citep{baltruvsaitis2018multimodal} to combine the information embedded in a borrower's loan application sequence and her demographics. The loan application sequence is processed by the transformer module into a feature vector $\mathbf{f}_A$, and the demographic information is processed by a feed-forward module into the feature vector $\mathbf{f}_D$. We then combine the two parts of information through element-wise addition $\mathbf{f} = \mathbf{f}_A \oplus \mathbf{f}_D$. And $\mathbf{f}$ is further normalized to be on a hypersphere.


\subsection{Contrastive Learning}

In order to learn from the unlabeled (unapproved/rejected) loan application records, we leverage contrastive learning to train our feature extractor. Contrastive learning learns effective representations in a self-supervised way by pulling semantically similar items closer together and pushing dissimilar items farther apart \citep{hadsell2006dimensionality}. 

In order to do contrastive learning, we need a set of paired examples, where each pair consists of two semantically related items. We follow \cite{gao-etal-2021-simcse} to use independently sampled dropout masks to create positive pairs. The dropout mask is the major component of the dropout layer, where masked units are set to zero during training. Specifically, for a batch of loan applications $\{\mathbf{A}_i\}_{i=1}^M$ and the corresponding applicant demographics $\{\mathbf{D}_i\}_{i=1}^M$, we denote the feature vector $\mathbf{f}_i^z = f_\theta(\mathbf{A}_i, \mathbf{D}_i, z)$, where $f_\theta$ is the feature extractor of our model and $z$ is the random dropout mask. We input the same sample into the feature extractor twice and we get two different feature embeddings $\mathbf{f}_i^z, \mathbf{f}_i^{z'}$ with two different random dropout masks $z$ and $z'$. The two different feature embeddings are a positive pair whose semantics are similar to each other. We do not use common data augmentation techniques such as cropping a sequence, deleting or replacing sequence elements, because these discrete augmentations may hurt performance.

We follow the contrastive learning framework in \cite{chen2020simple} to use the in-batch negatives \citep{chen2017sampling}, where we treat all other pairs within the same batch as the negative samples for a given positive pair. Concretely, we randomly sample a mini-batch of $M$ unlabeled training samples. We encode each sample twice with different dropout masks, so that the batch contains $M$ positive pairs of feature embeddings. Given a positive pair of feature embedding vectors $\mathbf{f}_i^z$ and $\mathbf{f}_i^{z'}$, the rest $2(M-1)$ feature vectors are used as negative samples. The contrastive loss for a batch is:
\begin{equation}
\scriptsize
\begin{aligned}
    \mathcal{L}_{\text{CL}} = -\frac{1}{2M} \sum_{i=1}^M  
     {\big[} \log \frac{\exp (\operatorname{sim}(\mathbf{f}_i^{z_i}, \mathbf{f}_i^{{z}_{i}'}) / \tau)}{\sum_{k=1, k\ne i}^{M} \exp (\operatorname{sim}(\mathbf{f}_i^{z_i}, \mathbf{f}_k^{z_k}) / \tau) + \sum_{k=1}^{M} \exp (\operatorname{sim}(\mathbf{f}_i^{z_i}, \mathbf{f}_k^{{z}_{k}'}) / \tau )} + \\
     \log \frac{\exp (\operatorname{sim}(\mathbf{f}_i^{{z}_{i}'}, \mathbf{f}_i^{z_i}) / \tau)}{\sum_{k=1}^{M} \exp (\operatorname{sim}(\mathbf{f}_i^{{z}_{i}'}, \mathbf{f}_k^{z_k}) / \tau) + \sum_{k=1, k\ne i}^{M} \exp (\operatorname{sim}(\mathbf{f}_i^{{z}_{i}'}, \mathbf{f}_k^{{z}_{k}'}) / \tau )} {\big]}
\end{aligned}
\end{equation}

\noindent where $\operatorname{sim}(\mathbf{f}_i, \mathbf{f}_j)$ is the cosine similarity $\operatorname{sim}(\mathbf{f}_i, \mathbf{f}_j)=\frac{\mathbf{f}_i^\mathsf{T} \mathbf{f}_j}{\| \mathbf{f}_i \| \cdot \|\mathbf{f}_j\|}$, and $\tau$ is a temperature hyperparameter. We set $\tau=0.1$ for our training.



\subsection{Domain Adaptation}

Although we have used contrastive learning to help the feature extractor learn effective representations of unlabeled samples, the label predictor is only trained on labeled ones, which may not generalize well to unlabeled ones. Therefore, in addition to using contrastive learning to train on the unlabeled dataset, we also introduce an unsupervised domain adaptation method by \citet{ganin2015unsupervised} to leverage a gradient reversal layer (GRL) to solve the distribution shift problem between the labeled data and the unlabeled ones. 
\if\kdd
Unsupervised domain adaptation is a machine learning technique that aims to improve the performance of a model on a target domain, where only unlabeled data is available, by leveraging the knowledge from a source domain where labeled data is available. 

The goal of unsupervised domain adaptation is to adapt the model trained on the source domain to perform well on the target domain, even though the target domain may be different from the source domain in terms of data distribution, feature representation, and other factors. This is achieved by aligning the feature representations of the source and target domains, so that the model trained on the source domain can effectively generalize to the target domain.
\fi

In our setting, our goal is to be able to predict the labels of data points from both the labeled dataset distribution $\mathcal{S}$ (source domain) and the unlabeled dataset distribution $\mathcal{T}$ (target domain). At training time, we have access to the training samples from both the labeled source domain $\mathcal{S}(x,y)$ and the unlabeled target domain $\mathcal{T}(x)$.
Our feature extractor outputs the encoded feature vector $\mathbf{f}$ for each sample. The feature vector is mapped into labels by a label predictor. 
We also use a domain classifier to map the feature vector $\mathbf{f}$ into the domain label $d$ -- which domain $\mathbf{f}$ comes from, $\mathcal{S}$ (the labeled domain) or $\mathcal{T}$ (the unlabeled domain). 

During training, we aim to optimize the feature extractor and the label predictor to minimize the label prediction loss on the labeled part of the training set ($\mathcal{S}$). This ensures the discriminative power of the features and good prediction performance on the labeled (source) domain. At the same time, we want to make the distributions of the features for the source domain ($\mathcal{S}$) and the target domain ($\mathcal{T}$) to be close to each other so that our model can also perform well on the unlabeled domain.

However, it is difficult to measure the dissimilarity between the $\mathbf{f}_\mathcal{S}$ and $\mathbf{f}_\mathcal{T}$ distributions because the feature space is high-dimensional and the distribution of features is continually evolving during the training process. We follow \cite{ganin2015unsupervised} to estimate this dissimilarity by examining the loss of the domain classifier, given that the domain classifier parameters have been trained to optimize the discrimination between the two feature distributions. Therefore, in addition to minimizing the loss of the label prediction, we simultaneously optimize the feature extractor parameters to maximize the loss of the domain classifier, and optimize the domain classifier parameters that minimize the loss of the domain classifier. The intuition here is the same as adversarial learning, where the domain classifier is a discriminator aiming to identify the feature vector's domain affiliation, while the feature extractor aims to generate features that are domain-invariant.

Mathematically, we realize this through the gradient reversal layer, which we denote as a pseudo-function $R(x)$. The forward and backward propagation of the gradient reversal layer is as follows:
\begin{align}
    &\text{Forward propagation: } R(\mathbf{x}) = \mathbf{x} \\
    &\text{Backward propagation: } \frac{\mathrm{d} R(\mathbf{x})}{\mathrm{d} \mathbf{x}} = -\lambda \mathbf{I}
\end{align}
where $\mathbf{I}$ is an identity matrix. The gradient reversal layer enables us to achieve our goal by directly minimizing the domain classification cross-entropy loss.

\section{Experiments}


\subsection{Experimental Setup}

For our experiments, all hidden dimensions are set to 64. We use a training batch size of 1024, and train our model for 15 epochs. The optimizer used is Adam with a learning rate of 0.001 and beta values of $\beta_1=0.9$ and $\beta_2=0.999$. The temperature hyperparameter $\tau$ for the contrastive loss is 0.1. And we set the loss weight $w_y$ to b1 1, $w_\text{CL}$ to be 0.1, and $w_d$ to be $0.1 \cdot (\frac{2}{1+\exp(-\gamma \cdot p)}-1)$, where $\gamma=0.001$.

\subsection{Performance Comparison}


Table~\ref{tab:models-auc-prof} presents the loan screening performance on the test dataset of our model and the ones without contrastive learning and/or domain adaptation. The first column indicates the name of each model. The second column shows a metric called AUC (Area Under The Curve) ROC (Receiver Operating Characteristics) curve.
\if\kdd
, which is a measure of the prediction performance on loan default. The higher the AUCROC, the better the classifier is at distinguishing default loans and non-default loans. 
\fi
The third column gives the profits generated by each model's prediction. 

\begin{table}[]
\centering
\begin{tabular}{lrr}
\toprule
\textbf{Model}                                                           & \textbf{AUCROC} & \textbf{Profits} \\ \midrule
Ours                                                                     & 0.7056      & 447564.97         \\
Ours w/o Contrastive Learning & 0.6723      & 423719.30         \\
Ours w/o Domain adaptation      & 0.6904      & 433388.89         \\
Ours w/o Both                                                            & 0.6588      & 410775.50         \\ \bottomrule
\end{tabular}
\caption{The AUC and profits of different models}
\label{tab:models-auc-prof}
\vskip -6mm
\end{table}

Based on the values in the table, our model achieves the highest AUCROC of 0.7056. The next highest AUCROC is for the model with just contrastive learning, which has an AUCROC of 0.6904. 
\if\kdd
This supports the effectiveness of the domain adaptation method in mitigating the distribution shift problem between approved loans and unapproved loans.
\fi
The model with just domain adaptation achieves an AUCROC of 0.6723, which is also lower than that of our model.   The vanilla model without neither contrastive learning nor domain adaptation has the lowest AUCROC, at 0.6588. These results suggest that the contrastive learning and the domain adaptation are effective at improving the loan screening performance by 4.80\% $(\frac{0.6904-0.6588}{0.6588})$ and 2.05\% $(\frac{0.6723-0.6588}{0.6588})$ respectively.

In terms of profits, our model generates the highest profits, at 447564.97. The model with just contrastive learning generates the next highest profits, at 433388.89. The model with just domain adaptation generates 423719.30 in profits. And vanilla model generates the lowest profits, at 410775.50. These indicate that the introduction of the contrastive learning and the domain adaptation yields economic gains of 5.51\% $(\frac{433388.89-410775.50}{410775.50})$ and 3.15\% $(\frac{423719.30-410775.50}{410775.50})$ in the platform profit respectively.
These performance improvements suggest that contrastive learning and domain adaptation are useful techniques for addressing the representation bias and the distribution shift problem. And they improve our model's ability to distinguish between default loans and non-default loans for the borrower population.

\subsection{Financial Inclusion Analysis}

Table~\ref{tab:inclusion} compares the characteristics of borrowers who are granted loans by different models. The rows represent different features or characteristics of the loan borrowers, including the living city DPI, the monthly income level of the borrower, the education level of the borrower, and the homeownership status of the borrower. The columns represent different versions of the model being used.
\if\kdd
, including our model, the model without contrastive learning, the model without domain adaptation, and the model without either contrastive learning or domain adaptation. 
\fi
The values in the table represent the mean of the feature for borrowers whose loan applications are approved by each version of the model.

\begin{table}[]
\centering
\footnotesize
\begin{tabular}{lrrrr}
\toprule
\textbf{}                                                                          \textbf{Feature}     & \multicolumn{1}{l}{\textbf{Ours}} & \multicolumn{1}{l}{\textbf{\begin{tabular}[c]{@{}l@{}}Ours w/o \\ Contrastive\\ Learning\end{tabular}}} & \multicolumn{1}{l}{\textbf{\begin{tabular}[c]{@{}l@{}}Ours w/o\\ Domain\\ adaptation\end{tabular}}} & \multicolumn{1}{l}{\textbf{\begin{tabular}[c]{@{}l@{}}Ours w/o \\ Both\end{tabular}}} \\ \midrule
 Living-city DPI      & 43912.90                          & 44835.41                                                                                                & 44559.26                                                                                          & 45495.44                                                                              \\
                                                                                                   Monthly income level & 3.9685                              & 4.2033                                                                                                    & 4.1274                                                                                              & 4.2682                                                                                  \\
                                                                                                   Education level      & 2.4615                              & 2.4819                                                                                                    & 2.4626                                                                                              & 2.5060                                                                                  \\
                                                                                                   Homeownership        & 0.1983                              & 0.2070                                                                                                    & 0.2054                                                                                              & 0.2177                                                                                  \\ \bottomrule
\end{tabular}
\caption{Comparison of the characteristics of the loans approved by different models}
\label{tab:inclusion}
\vskip -6mm
\end{table}

The model with contrastive learning and domain adaptation has the lowest mean values for the four demographic features, which indicates that this model is more inclusive in approving loans to borrowers with lower socioeconomic backgrounds. Comparing the specific numeric values in the table, we can see that the mean living city DPI is lower for our model that uses both contrastive learning and domain adaptation (43912.90) compared to the model with just domain adaptation (44835.41), the model with just contrastive learning (44559.26), and the model that uses neither technique (45495.44). Similarly, the mean monthly income level is lower for our model (3.9685) compared to the model with just domain adaptation (4.2033), the model with just contrastive learning (4.1274), and the model that uses neither technique (4.2682). Similarly, the mean education level and the mean homeownership of approved borrowers also decrease with the use of contrastive learning and domain adaptation.
\if\kdd
These results indicate that the use of contrastive learning and domain adaptation results in lower feature means for the approved loan borrowers. These techniques are allowing the model to consider a broader range of borrowers, including those from lower socioeconomic backgrounds. The model with both contrastive learning and domain adaptation tends to approve loans to borrowers with lower living city DPI, lower monthly incomes, lower education levels, and worse home ownership. And the use of both contrastive learning and domain adaptation is more effective at achieving financial inclusion than using just one of these techniques.
\fi

The improvements in financial inclusion benefit from domain adaptation and contrastive learning enabling the model to mitigate the distribution shift problem and to learn from a more diverse set of unlabeled data, which contains information about borrowers with different socioeconomic characteristics. This increased diversity in the training data may help the model to better capture the complexity and nuance of real-world whole borrower pool distributions, leading to more fair and equitable lending decisions.

\if\kdd
{\color{red} TODO: apprIdx}
\fi


\subsection{Performance on Different Sequence Lengths}

\begin{figure}[]
\centering
\begin{center}
\includegraphics[width=0.4\textwidth]{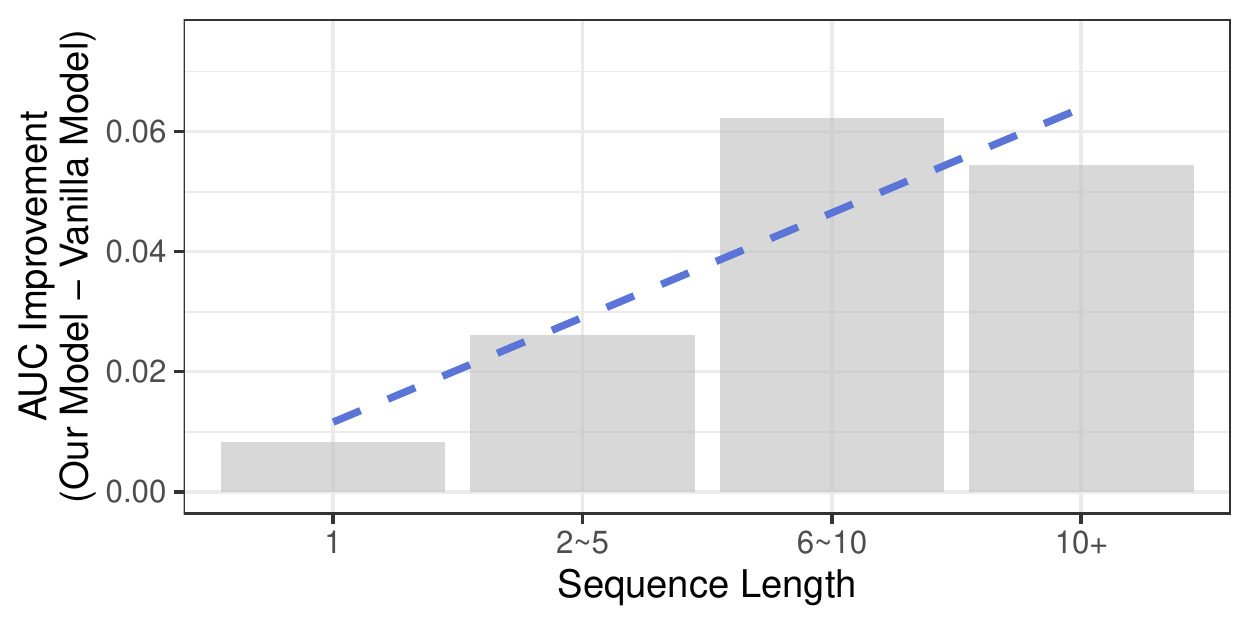}
\vskip -2mm
\caption{AUC performance improvement for different sequence lengths} 
\label{fig:auc-length}
\end{center}
\vskip -6mm
\end{figure}

An important module of our model is the transformer module for sequential modeling. To investigate our model's performance on loan sequences of different lengths, we plot the AUCROC performance improvement of our model relative to the vanilla model for different sequence lengths (Figure~\ref{fig:auc-length}). The x-axis is divided into four bins based on the length of the loan application sequences of the users in the test dataset, with the first group only having loan applications once and the fifth group having applied more than 10 times. The y-axis represents the difference in performance between our model and the vanilla model, which does not use contrastive learning or domain adaptation. As the figure shows, our model performs better than the vanilla model for all different sequence lengths. The performance of our model is found to improve more in longer sequence groups, as indicated by the positive slopes of the fitted line (the dotted line) in Figure~\ref{fig:auc-length}. These results suggest that the sequence embedding approach of our model can effectively learn semantics from long sequences to benefit data augmentation, contrastive learning, and domain adaptation.


\subsection{PCA Projections of the Embeddings}

\begin{figure*}[]
  \begin{subfigure}{0.245\textwidth}
    \includegraphics[width=\linewidth]{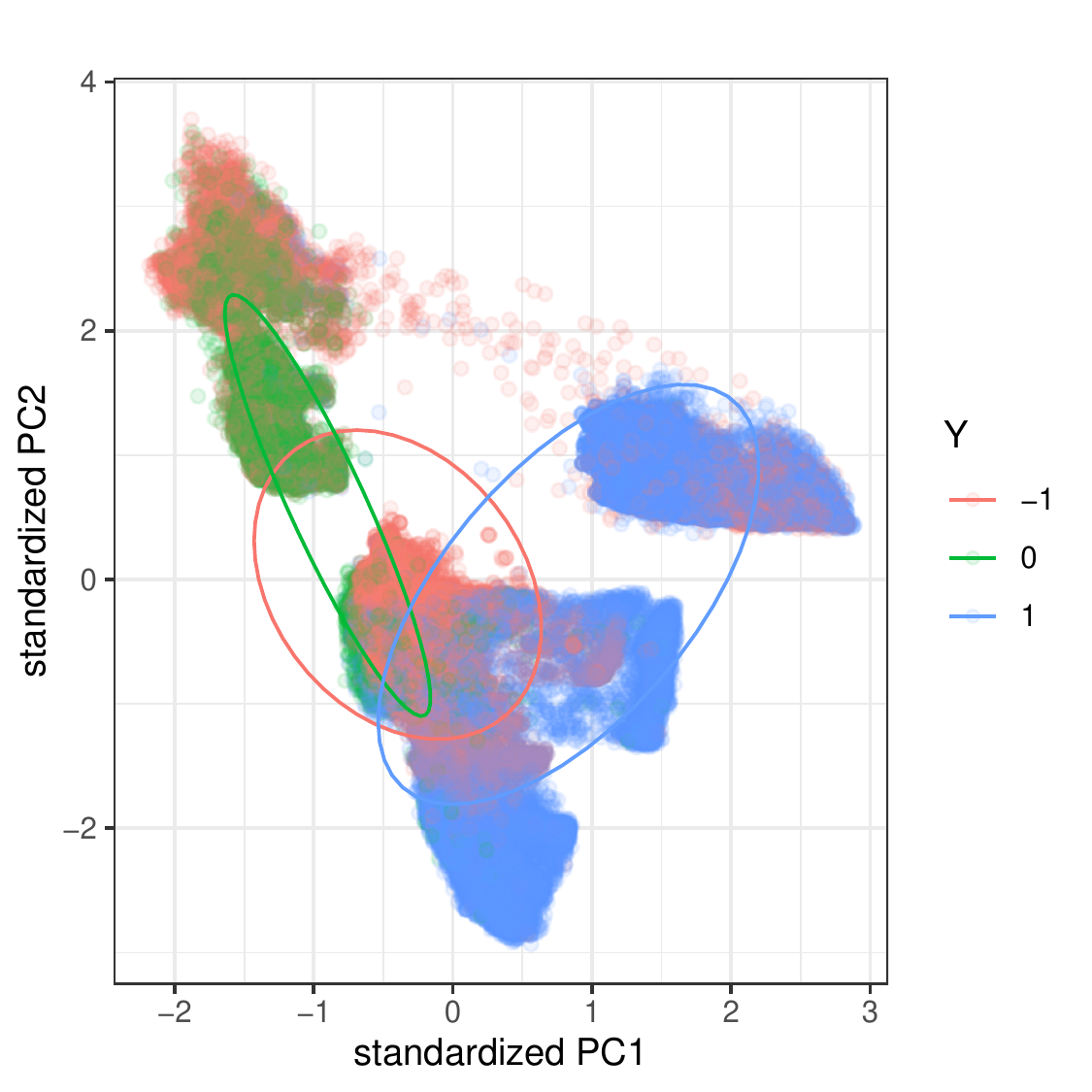}
    \caption{Vanilla Model}
    \label{fig:pca-vanilla}
  \end{subfigure}%
  \begin{subfigure}{0.245\textwidth}
    \includegraphics[width=\linewidth]{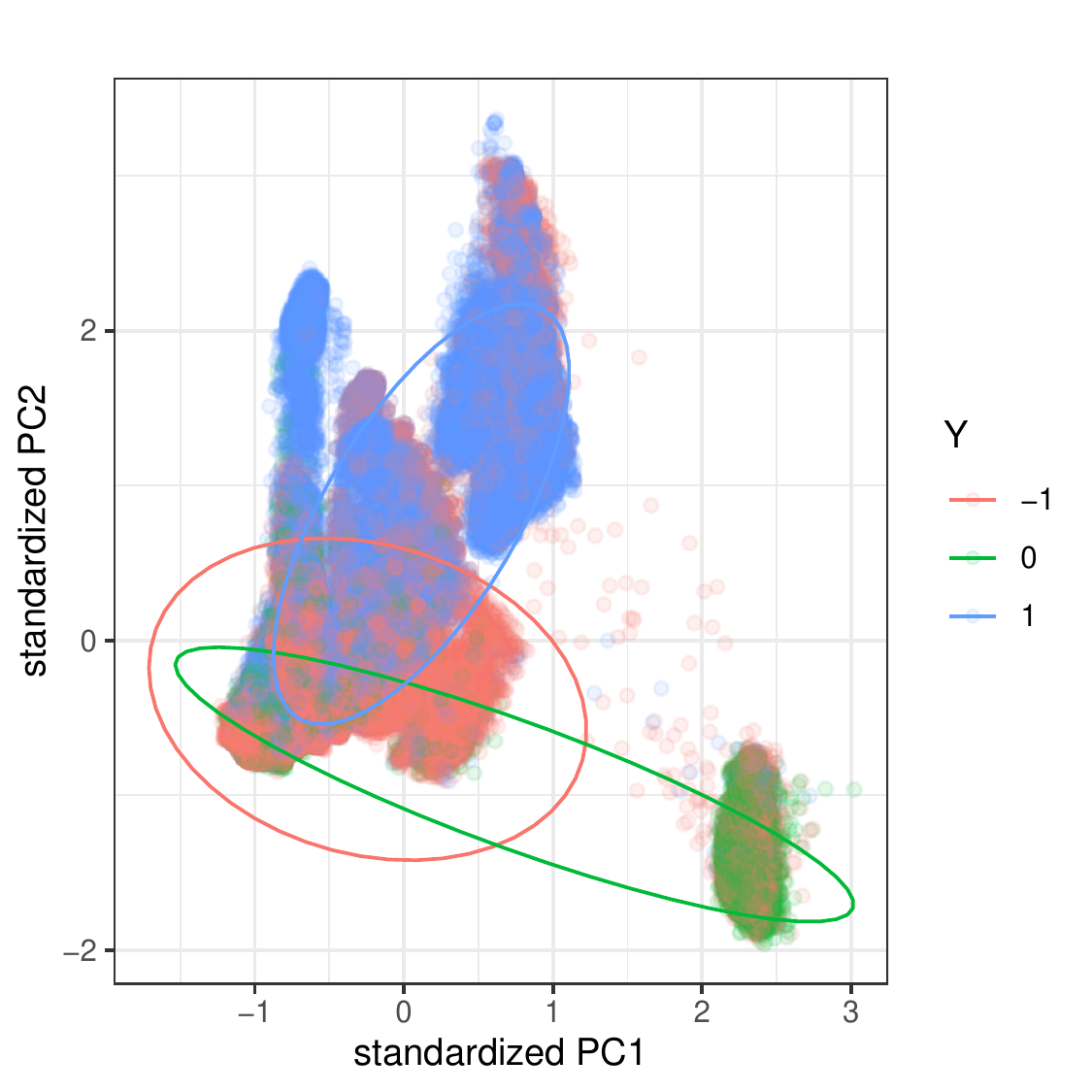}
    \caption{model w/o contrastive learning}
    \label{fig:pca-da}
  \end{subfigure}
  \begin{subfigure}{0.245\textwidth}
    \includegraphics[width=\linewidth]{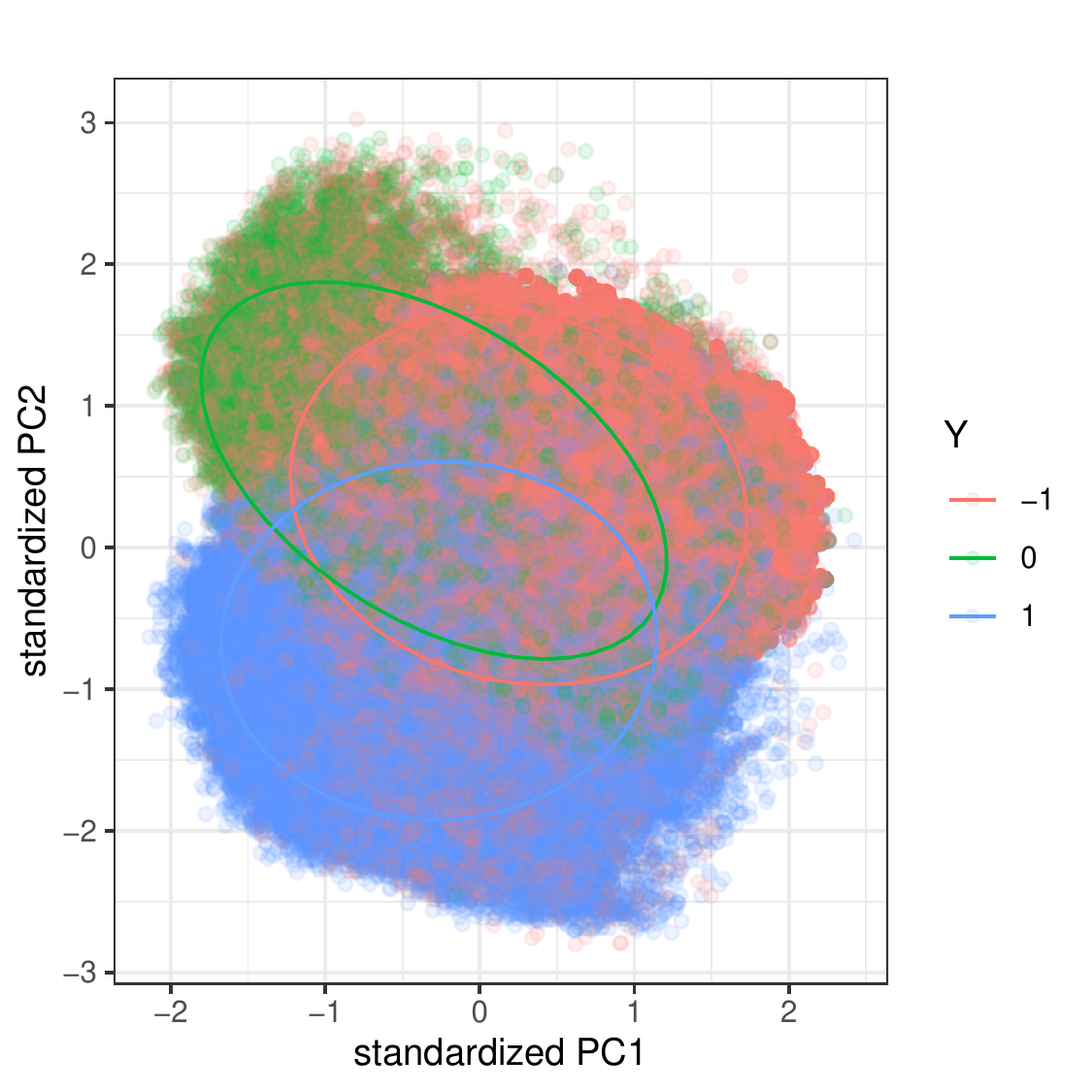}
    \caption{model w/o domain adaptation}
    \label{fig:pca-cl}
  \end{subfigure}%
  \begin{subfigure}{0.245\textwidth}
    \includegraphics[width=\linewidth]{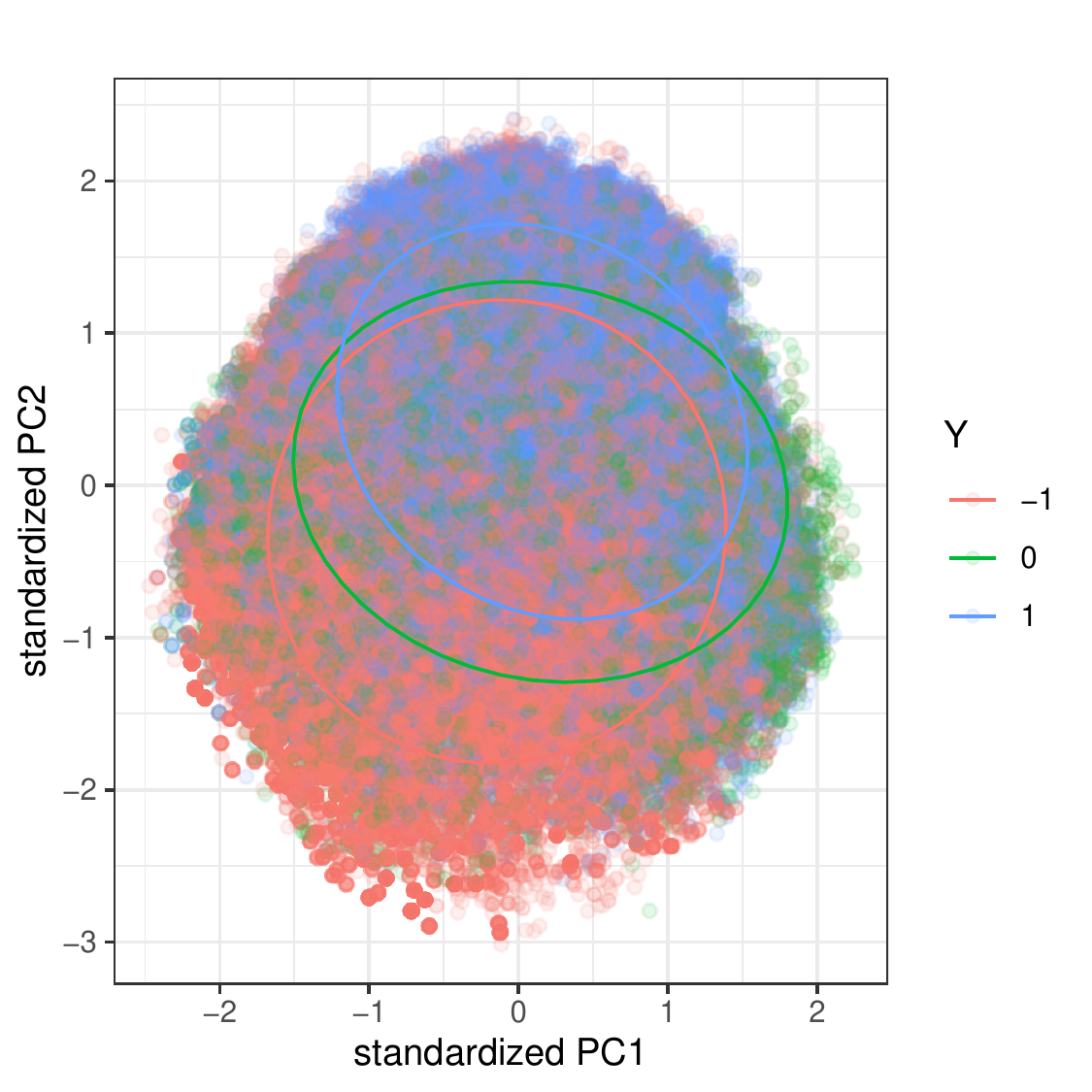}
    \caption{Our model}
    \label{fig:pca-cl-da}
  \end{subfigure}
  \caption{Visualization of the training data feature representations $\mathbf{f}$ of different models. The dimension of these representations is reduced to 2 by Principle Component Analysis (PCA). Blue points are loans being approved and non-default. Green points are loans being approved and default. Red points are loans not being approved.}
  \label{fig:pca}
\end{figure*}

To investigate the feature embeddings learned by different models, we use the PCA projection to visualize the distribution of feature vectors, and we color-code the labels and domains (as shown in Figure~\ref{fig:pca}). We find that the success in terms of loan screening accuracy for the test dataset is strongly correlated with the overlap between the domain distributions in these visualizations.

Figure~\ref{fig:pca} demonstrates the effect of domain adaptation and contrastive learning on the distribution of extracted feature vectors of the training dataset. The samples are colored by the label $Y$ (1 is non-default, 0 is default, and -1 is unlabeled). It shows PCA visualizations of our models with and without domain adaptation and/or contrastive learning. In the visualization, blue points represent loan examples that are approved and non-default, green points represent loan examples that are approved and default, and red points represent examples from unapproved loans whose labels are unobserved.

For the vanilla model (Figure~\ref{fig:pca-vanilla}), most of the unlabeled data points (red) overlap with the default data points (green). This makes the classifier tend to underestimate the creditworthiness of people from low socioeconomic backgrounds. The domain adaptation (Figure~\ref{fig:pca-da}) pushes the distribution of the unlabeled data points closer to that of the non-default ones. The contrastive learning (Figure~\ref{fig:pca-cl}) improves the uniformity of the distribution and pushes all the distributions of the three sets of samples closer to each other. When we use both the contrastive learning and the domain adaptations, our approach aligns the feature distributions well and keeps appropriate distinguishability among the three classes (as shown in Figure~\ref{fig:pca-cl-da}), which results in successful adaptation and classification performance.

\subsection{Alignment and Uniformity of Embeddings}

In contrastive learning, a model is trained to identify positive and negative examples in a dataset and to maximize the distance between the positive and negative examples in the learned representation space. We use two key properties identified by \cite{wang2020understanding}, i.e. \textit{alignment} and \textit{uniformity}, to measure the quality of learned representations. Alignment refers to the expected distance between the embeddings of paired instances in the learned representation space, while uniformity refers to the degree to which the examples are evenly distributed on the hypersphere. These two metrics align with the goal of contrastive learning, which is to have embeddings for paired instances remain close together and to have embeddings for random instances scattered on the hypersphere. The calculation of alignment and uniformity is as follows:
\begin{align}
    &\ell_{\text{align}} \equiv \underset{x \sim p_\text{data}}{\mathbb{E}} \left\| f_\theta(x, z) - f_\theta(x, z') \right\|^2 \\
    &\ell_{\text{uniform}} \equiv \underset{x,y \overset{i.i.d.}{\sim} p_\text{data}}{\mathbb{E}} e^{-2\left\| f_\theta(x) - f_\theta(y) \right\|^2}
\end{align}
where $p_\text{data}$ denotes the data distribution, $f_\theta$ is the feature extractor of our model, and $z$ is the random dropout mask.

\begin{figure}[]
\centering
\begin{center}
\includegraphics[width=0.4\textwidth]{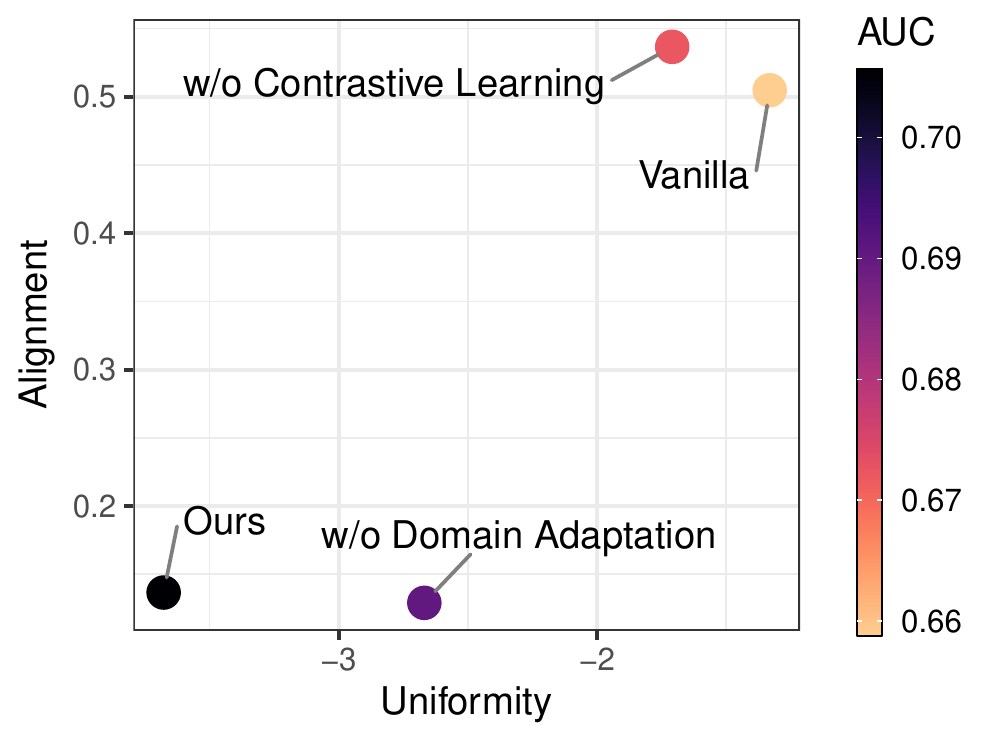}
\vskip -5mm
\caption{The alignment and uniformity of different models (colored by AUC)} 
\label{fig:align-uniform}
\end{center}
\vskip -4mm
\end{figure}

Figure~\ref{fig:align-uniform} shows the alignment and the uniformity of various loan embedding models, along with their AUC performance. We find that models with both good alignment and uniformity (points on the bottom left) tend to perform better, which is consistent with the findings of \cite{wang2020understanding}. We notice that contrastive learning improves both the alignment and the uniformity of embeddings. We also notice that domain adaptation has better uniformity but worse alignment. With both the contrastive learning and the domain adaptation, we achieve the best uniformity and sacrifice very little alignment compared with using contrastive learning without domain adaptation.

\subsection{Replace Transformer with Other Backbones (RNN, GRU, LSTM)}

To test the robustness of our method, we replace the Transformer module with other sequential neural networks. Table~\ref{tab:backbones-auc-prof} presents the results of a study that compares the performance of three different models based on three different backbones: RNN, LSTM, and GRU. The models are evaluated based on two metrics: AUCROC and profits. The table shows that for all three backbones, the model that includes both contrastive learning and domain adaptation (Ours) performs the best in terms of both AUCROC and profits. When contrastive learning or domain adaptation is removed from the model, the performance decreases for both metrics. When both contrastive learning and domain adaptation are removed, the performance decreases even further. This suggests that the performance gain of introducing contrastive learning and domain adaptation is consistent across the three different sequential neural network backbones.

\begin{table}[]
\centering
\begin{tabular}{@{}lrr@{}}
\toprule
\textbf{Model}                & \textbf{AUCROC}      & \textbf{Profits}     \\ \midrule
\textit{RNN}                  & \multicolumn{1}{l}{} & \multicolumn{1}{l}{} \\
Ours                          & 0.6899              & 422813.54            \\
Ours w/o Contrastive Learning & 0.6532              & 392086.35            \\
Ours w/o Domain Adaptation    & 0.6688              & 408805.38            \\
Ours w/o Both                 & 0.6309              & 389354.77            \\ \midrule
\textit{LSTM}                 & \multicolumn{1}{l}{} & \multicolumn{1}{l}{} \\
Ours                          & 0.7094              & 438939.45            \\
Ours w/o Contrastive Learning & 0.6714              & 422891.53            \\
Ours w/o Domain Adaptation    & 0.6857              & 429204.09            \\
Ours w/o Both                 & 0.6491              & 400217.83            \\ \midrule
\textit{GRU}                  & \multicolumn{1}{l}{} & \multicolumn{1}{l}{} \\
Ours                          & 0.7032              & 447725.81            \\
Ours w/o Contrastive Learning & 0.6692              & 420226.36            \\
Ours w/o Domain Adaptation    & 0.6871              & 431375.42            \\
Ours w/o Both                 & 0.6557              & 405733.56            \\ \bottomrule
\end{tabular}
\caption{The AUC and profits of different models based on different backbones (RNN, LSTM, GRU)}
\label{tab:backbones-auc-prof}
\vskip -6mm
\end{table}

\subsection{Incorporate Test Dataset in an Unlabeled Way}

All our experiments above only involve the unlabeled training samples into the domain adaptation and the contrastive learning losses. Since the two modules do not require any labels, we can also incorporate the test samples. This can mitigate the potential problem that the overall borrowers' distribution on the market may keep changing over time---the test samples shift from training samples. 

Table~\ref{tab:testdata-auc-prof} presents the results of a study that compares the performance of two different models. The models are evaluated based on two metrics: AUCROC and profits. According to the table, the model labeled "Ours" (which includes contrastive learning and domain adaptation on unlabeled training samples) achieves an AUCROC of 0.70564 and generates profits of 447564.97. The model labeled "Ours + also use test data unsupervisedly in CL and DA" (which includes the same contrastive learning and domain adaptation as the first model, but also uses test data unsupervisedly in these processes) achieves an AUCROC of 0.71416 and generates profits of 458685.60. This suggests that incorporating the test samples in the unsupervised domain adaptation and contrastive learning improves the loan screening performance by 1.20\% $(\frac{0.7141-0.7056}{0.7056})$ and the economic gain by 2.48\% $(\frac{458685.60 - 447564.97}{447564.97})$.

\begin{table}[]
\centering
\begin{tabular}{lrr}
\toprule
\textbf{Model}                                                           & \textbf{AUCROC} & \textbf{Profits} \\ \midrule
Ours                                                                     & 0.7056      & 447564.97         \\
Ours + use test data in CL and DA &	0.7141 & 458685.60         \\ \bottomrule
\end{tabular}
\caption{The AUC and profits of different models}
\label{tab:testdata-auc-prof}
\vskip -6mm
\end{table}

\subsection{Effects of Label Some Test Samples}

In the real world, using semi-supervised or weakly-supervised methods is often preferable, as it can be difficult or time-consuming to obtain a large amount of labeled data. In many cases, obtaining even a small ratio of labeled data is more feasible to boost model performance at a low cost. In this section, we test the model's performance when we randomly approve a small ratio of test samples to obtain their labels and add them to the labeled training dataset.

\begin{figure}[]
  \begin{subfigure}{0.24\textwidth}
    \includegraphics[trim=1mm 0mm 0mm 0mm, clip, width=0.97\textwidth]{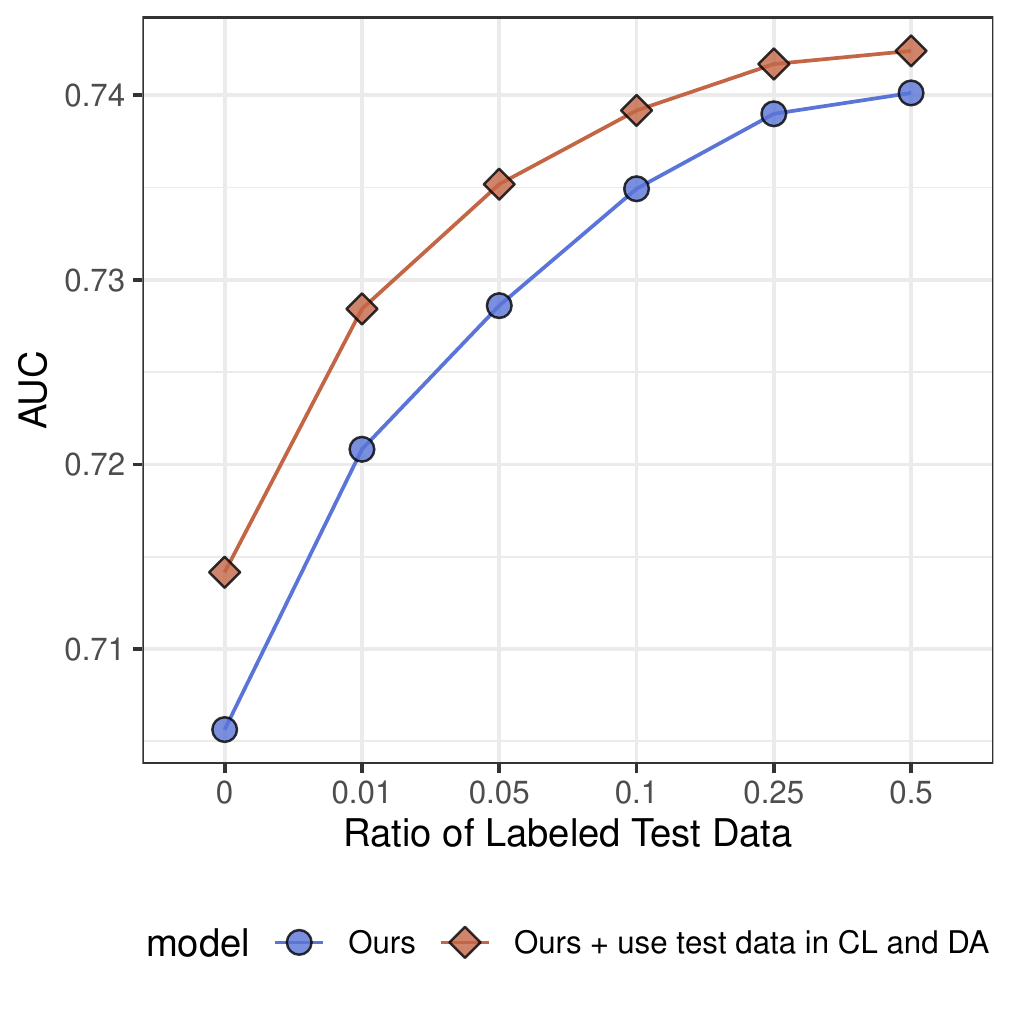}
    \vskip -1.8mm
    \caption{AUC} 
    \label{fig:labeled-ratio-auc}
  \end{subfigure}%
  \begin{subfigure}{0.245\textwidth}
    \includegraphics[width=0.98\textwidth]{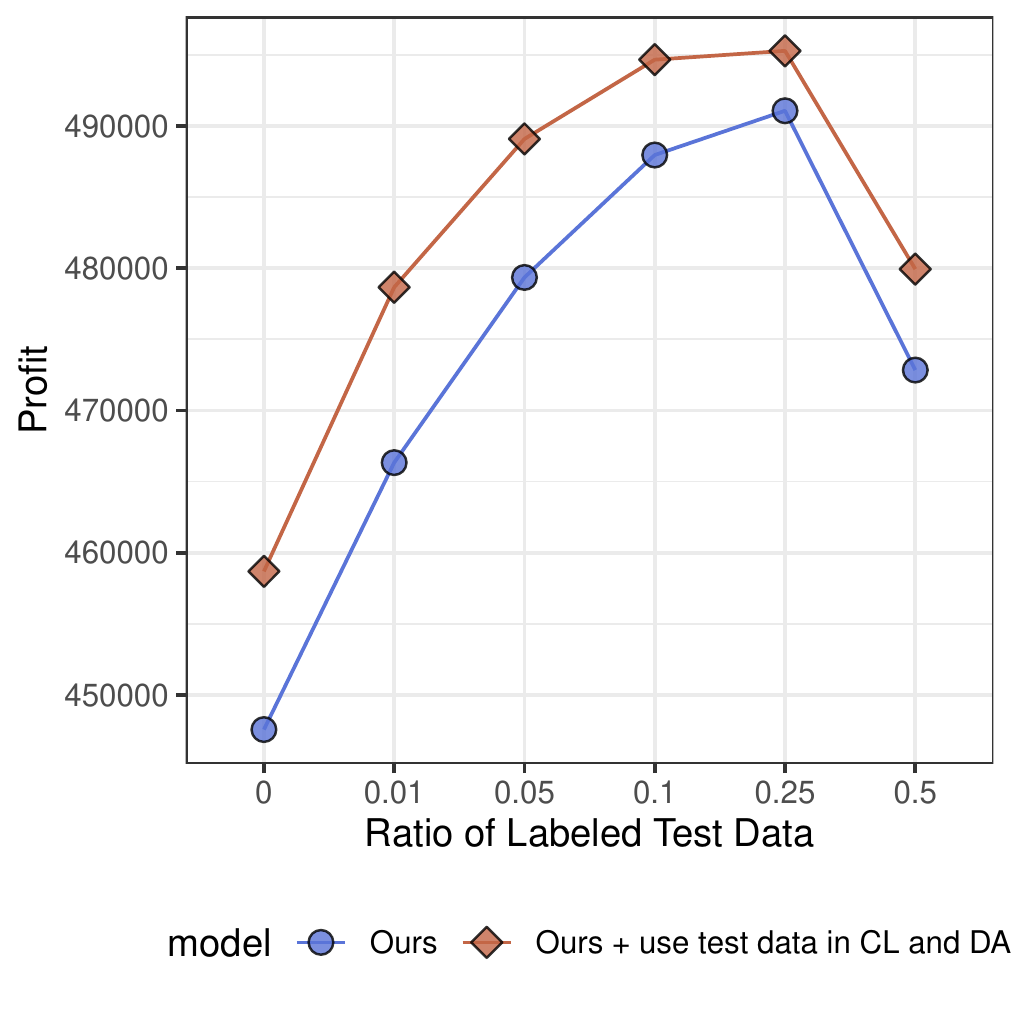}
    \vskip -2mm
    \caption{Profit} 
    \label{fig:labeled-ratio-profit}
  \end{subfigure}
  \caption{AUC and profit of different ratios of labeled test samples}
  \label{fig:labeled-ratio}
  \vskip -7mm
\end{figure}

In Figure~\ref{fig:labeled-ratio}, we plot the model performance in terms of their AUCROC and profits when using different ratios of labeled test data. We experiment on our models with and without using the unlabeled test data in the domain adaptation and the contrastive learning. We test different proportions of labeled test data used, ranging from 0 (no labeled test data) to 0.5 ("randomly approve" half of the loan applications in the test data to get their labels). For the AUCROC (Figure~\ref{fig:labeled-ratio-auc}), we only calculate the model's performance on the rest unlabeled test samples. We observe that the model performance increases as the ratio of labeled test data increases for both models. We observe that even just labeling 1\% of the test data can lead to a significant improvement in model performance. Our model achieves an AUCROC of 0.7056 when no labeled test data was used, but when 1\% of the test data was labeled, the AUCROC increased to 0.7208. Similarly, when our model incorporates unlabeled test samples in domain adaptation and contrastive learning, it has an AUCROC of 0.7141 with no labeled test data, but an AUCROC of 0.7284 when 1\% of the test data was labeled. This suggests that even a small amount of labeled test data can have a significant improvement on model performance.

We also note that using unlabeled test data in contrastive learning and domain adaptation consistently performs better than the model not using it, as indicated by the higher AUCROC values in the figure. However, the gap between the two models' performance decreases as the labeled test data ratio increases. For example, the difference between the two model's AUCROC is about 0.0085. At a labeling ratio of 0.5, the difference is just about 0.0022. This suggests that the benefit of using unlabeled test data in contrastive learning and domain adaptation decreases as the amount of labeled test data increases.


For the profits (Figure~\ref{fig:labeled-ratio-profit}), we calculate the profits of all the test samples, including (1) the profits from all the randomly approved test samples, and (2) the profits from all the rest test samples which are screened by the trained model. Overall, the profit increases as the ratio of labeled test data increase for both models. But we also note that labeling (randomly approving) too much test data could lead to a decrease in profit. This occurs when randomly approving the additional loan applications results in funding low-quality borrowers with negative profits, which outweigh the gains of the improved model performance on the rest test samples. It would be important to carefully consider the trade-off between the cost of labeling additional data and the expected increase in profit when deciding on the amount of labeled test data to use.

\section{Related Literature}

\subsection{Machine Learning for Financial Risk}

Machine learning techniques have been widely used in financial risk modeling \cite{ahelegbey2019latent, chen2020encoding}. Studies such as \cite{guo2008neural, puh2019detecting, nur2020comparative, liu2022bond, li2022ecod, han2022adbench, chen2022antibenford} proposed or applied advanced machine learning techniques to detect financial fraud. Additionally, research such as \cite{villuendas2017naive, xia2018novel, babaev2019rnn, liang2021credit} used machine learning for credit scoring. Other studies such as \cite{lee2007application} used machine learning methods for bond rating prediction, \cite{bi2022company} designed a graph neural network for company financial risk assessment, and \cite{yue2022multi} proposed a multi-task Transformer-based model to predict corporate credit rating migration. 
To tackle the bias problem in financial risk modeling, many researchers invent new machine learning algorithms by adding fairness constraints or incorporating fairness objectives to ensure adherence to fairness requirements in decision-making \cite{agarwal2018reductions, donini2018empirical, hu2022credit}. In addition, some work proposes to detect and mitigate algorithmic bias by enhancing algorithm transparency and interpretability \cite{rudin2019globally, rudin2019stop, hu2019optimal}.

\subsection{Self-supervised Learning}

Self-supervised learning is a method for training models without the need for labeled data. 
Mainstream methods of self-supervised learning can be broadly classified into three categories: Generative, Contrastive, and Generative-Contrastive (Adversarial) \cite{liu2021self}. The generative approach trains an autoencoder to learn feature representations. The contrastive approach trains an encoder to maximize the similarity between similar samples. The Generative-Contrastive approach trains an encoder-decoder to generate fake samples and a discriminator to distinguish the fake ones from real ones. In this work, we follow the contrastive learning framework of \cite{chen2020simple} to maximize the agreement between positive samples, and minimize the similarity between positive and negative ones.

\subsection{Domain Adaptation}

Domain adaptation is a technique used in machine learning to adapt a model trained on one dataset or domain to work effectively on a different but related dataset or domain. This is often done by minimizing the difference between the distributions of the source and target domains.
It has wide applications in many fields, including computer vision~\cite{tzeng2014deep,long2015learning},
and natural language processing~\cite{blitzer2007biographies,ganin2016domain}.
According to the amount of \textit{labeled data} in the target domain,
the domain adaptation task can be divided into three major categories---\textit{unsupervised}, \textit{semi-supervised}, and \textit{supervised} domain adaptation.
These three categories correspond to the cases where none, a few, and sufficient, labeled data in the target domain(s) are available.
Our work is closely related to unsupervised domain adaptation.
To address the discrepancy between feature distributions of data in the source and target domain unsupervisedly, various methods were proposed to learn domain invariant representations,
including using the Maximum Mean Discrepancy (MMD) loss~\cite{tzeng2014deep,long2015learning,sun2016deep},
minimizing domain shift using an adversarial loss~\cite{tzeng2015simultaneous,tzeng2017adversarial,ganin2016domain,li2017end,li2019transferable, ganin2015unsupervised},
or using a self-supervision loss~\cite{ghifary2015domain,ghifary2016deep,feng2019self,kang2019contrastive,li2020cross}.



\section{Conclusion}

In this work, we focus on the financial exclusion problem in FinTech lending caused by the representation bias and the distribution shift problems. We use Transformer to encode the loan sequence, and we propose to use self-supervised contrastive learning and unsupervised domain adaptation to address these problems by incorporating unlabeled loan application samples. Our experiment results suggest that our model outperforms baseline methods, and it can be used with a small fraction of labeled test data to further boost performance. Future work may extend our model to settings where nontraditional alternative data are available.


\bibliographystyle{ACM-Reference-Format}
\bibliography{refs}


\end{document}